\documentclass[sigconf]{acmart}

\usepackage{xspace}
\newcommand{\eigent}{\textsf{Eigent}\xspace}
\newcommand{\qplus}{\textsf{Q+}\xspace}
\newcommand{\eigentsearchqplus}{\textsf{EigentSearch-Q+}\xspace}

\usepackage{graphicx}
\usepackage{booktabs}
\usepackage{multirow}
\usepackage{xcolor}
\usepackage{hyperref}
\usepackage{cleveref}
\usepackage{listings}
\usepackage{comment}
\usepackage{placeins}
\usepackage{float}

\begin{document}
\acmYear{2026}\copyrightyear{2026}
\setcopyright{cc}
\setcctype[4.0]{by}
\acmConference[CAIS '26]{Proceedings of the 1st ACM Conference on Agentic and AI Systems}{May 26, 2026}{San Jose, CA, USA}
\acmBooktitle{Proceedings of the 1st ACM Conference on Agentic and AI Systems (CAIS '26), May 26, 2026, San Jose, CA, USA}
\acmDOI{10.1145/3786335.3813186}
\acmISBN{979-8-4007-2415-2/26/05}
\title{\eigentsearchqplus: Enhancing Deep Research Agents with Structured Reasoning Tools}

\author{Boer Zhang$^{1,*}$, Mingyan Wu$^{2,*}$, Dongzhuoran Zhou$^{3,4,*}$, Yuqicheng Zhu$^{5,4,*}$, Wendong Fan$^{6,7,*}$, Puzhen Zhang$^{6,7,*}$, Zifeng Ding$^{8,9,*}$, Guohao Li$^{6,7,*}$, Yuan He\textsuperscript{10,*,\textdagger}}
\affiliation{
  \institution{$^1$Meta, $^2$Northeastern University, China, $^3$University of Oslo, $^4$Bosch Center for AI, $^5$University of Stuttgart, $^6$CAMEL-AI.org, $^7$Eigent.AI, $^8$University of Cambridge, $^9$Mina AI, \textsuperscript{10}Amazon}
  \country{}
  \textsuperscript{*} This work was conducted while the authors were open-source contributors at CAMEL-AI.org.\\
  \textsuperscript{\textdagger} Corresponding author.
}
\renewcommand{\shortauthors}{}

\begin{abstract}
Deep research requires reasoning over web evidence to answer open-ended questions, and it is a core capability for AI agents. Yet many deep research agents still rely on implicit, unstructured search behavior that causes redundant exploration and brittle evidence aggregation.
Motivated by Anthropic's ``think'' tool paradigm and insights from the information-retrieval literature, we introduce \qplus, a set of query and evidence processing tools that make web search more deliberate by guiding query planning, monitoring search progress, and extracting evidence from long web snapshots.
We integrate \qplus into the browser sub-agent of \eigent, an open-source, production-ready multi-agent workforce for computer use, yielding \eigentsearchqplus.
Across four benchmarks (SimpleQA-Verified, FRAMES, WebWalkerQA, and XBench DeepSearch), \qplus improves \eigent's browser agent benchmark-size-weighted average accuracy by 3.0, 3.8, and 0.6 percentage points (pp) for GPT-4.1, GPT-5.1, and Minimax M2.5 model backends, respectively.
Case studies further suggest that \eigentsearchqplus produces more coherent tool-calling trajectories by making search progress and evidence handling explicit.\footnote{GitHub repository: \url{https://github.com/camel-ai/eigent_search}. Demo video: \url{https://youtu.be/Gea5esZllbE}.}
\end{abstract}

\ccsdesc[500]{Information systems~Information retrieval}
\ccsdesc[500]{Computing methodologies~Natural language processing}
\keywords{deep research, information retrieval, structured reasoning, large language models, AI agents}

\maketitle

\section{Introduction}
Deep research -- open-ended information seeking that requires iteratively searching, reading, and synthesizing evidence -- is an increasingly important application of large language model (LLM) agents~\cite{huang2025deepresearchagentssystematic}, and is now supported by a growing ecosystem of both proprietary systems (e.g., OpenAI Deep Research~\cite{openai2025introducingdeepresearch}, Gemini Deep Research~\cite{google2025geminideepresearch}, Grok DeepSearch~\cite{xai2025grokdeepsearch}, and Perplexity Deep Research~\cite{perplexity2025deepresearch}) and open-source applications (e.g., Search-o1~\cite{li2025searcho1agenticsearchenhancedlarge}, Search-R1~\cite{jin2025searchr1trainingllmsreason}, WebWalker~\cite{wu2025webwalkerbenchmarkingllmsweb}, AgentOrchestra~\cite{zhang2026agentorchestraorchestratingmultiagentintelligence}).
In this setting, an agent (or multi-agent system) is expected to interpret ambiguous or complex user queries, iteratively acquire and aggregate external information from heterogeneous sources (e.g., news, wikipages, social media), and decide when the collected evidence is sufficient to synthesize a comprehensive, grounded output.

While prior work has established the importance of planning and reasoning for deep-research agents, typical systems do not explicitly structure these processes at the tool level: reasoning is often left to the backend LLM, and planning is frequently implemented as a fixed workflow.
In this paper, we ask whether making planning and reasoning explicit via structured, tool-mediated thinking can improve the efficiency and robustness of deep-research agents.
Inspired by Anthropic's ``think'' tool paradigm~\cite{anthropic2025claude_think_tool} (which externalizes intermediate reasoning in a structured trace without invoking external capabilities), we develop \qplus: a suite of dedicated reasoning tools that make query and evidence processing operations explicit and inspectable.
Crucially, \qplus tools do not retrieve new external information; instead, they provide cognitive scaffolding by requiring the model to record both the \emph{inputs} and the \emph{expected intermediate outputs} of a reasoning step as typed tool arguments.
As a result, intermediate decisions become explicit, auditable, and directly linked to subsequent actions.
We integrate \qplus into an existing deep-research agent with two core capabilities:
(i) \emph{query processing} tools that, drawing on insights from information retrieval, formulate new queries, maintain a frontier of candidates, and prioritize promising directions for exploration; and
(ii) \emph{evidence processing} tools that extract relevant details from long web snapshots, and reflect on search progress to determine whether sufficient information has been gathered.

We demonstrate \qplus by integrating it into the browser agent of \eigent (~\cite{eigent2026website}), an open-source multi-agent workforce for computer use. In \eigent, deep-research behavior arises from system-level task decomposition and orchestration across multiple agents that focus on different tasks, while its browser agent is responsible for the web retrieval loop (searching and  browsing) that supports long-horizon investigation. We name the integrated agent system \eigentsearchqplus. Across four benchmarks (SimpleQA-Verified, FRAMES, WebWalkerQA, and XBench DeepSearch), we found \eigentsearchqplus improves the \eigent browser agent's accuracy by 3.0, 3.8, and 0.6 percentage points (pp) on average (weighted by benchmark size) when using GPT-4.1, GPT-5.1, and Minimax M2.5 model backends, respectively.
Beyond accuracy, case studies suggest that \qplus yields more structured trajectories with clearer query decomposition, more targeted extraction, and more explicit self-checks. More broadly, \qplus demonstrates a lightweight, non-invasive way to improve the robustness and observability of deep-research agents.

\begin{figure}[H]
  \centering
  \includegraphics[width=\columnwidth]{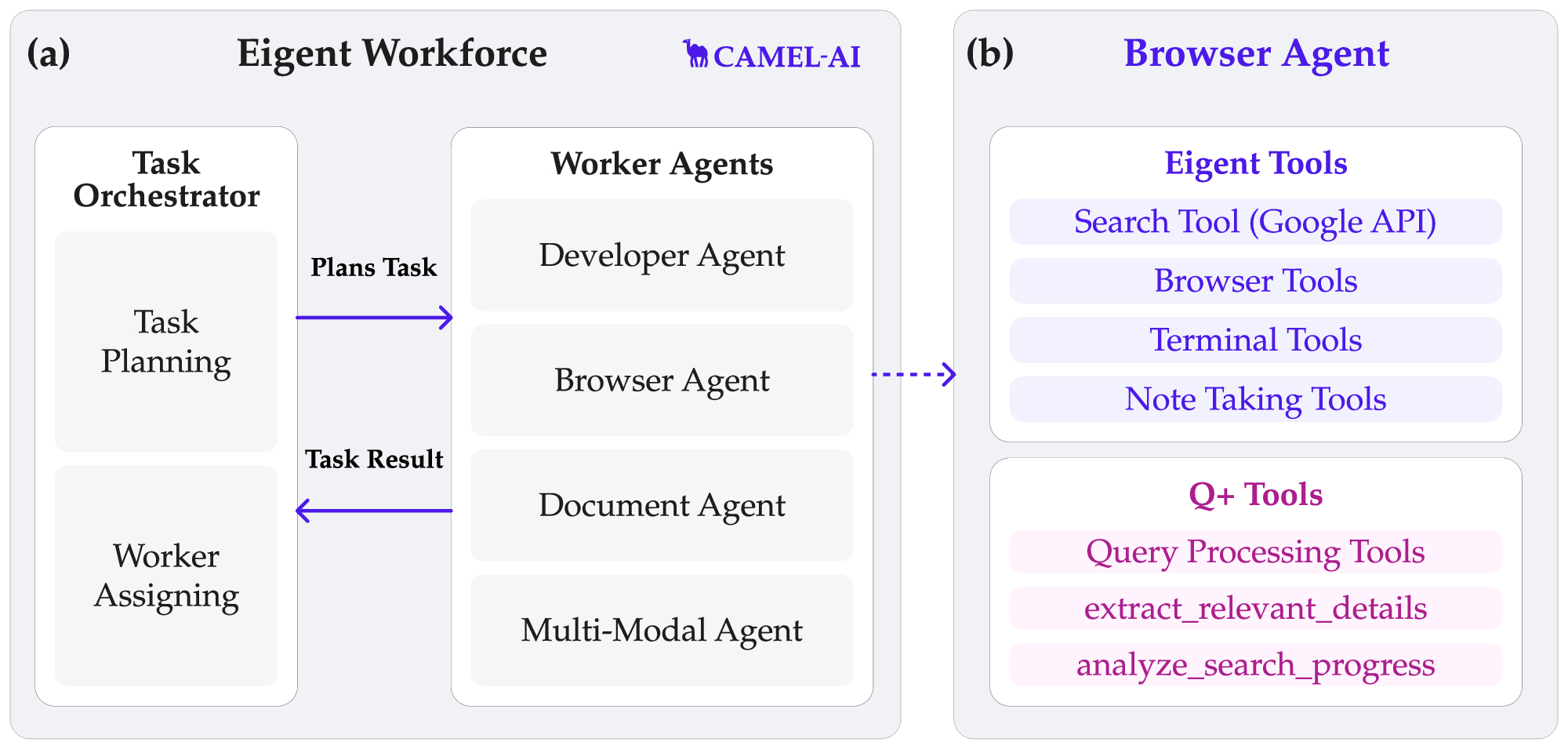}
  \caption{System architecture of the \eigent multi-agent framework and the \eigentsearchqplus enhancement. (a) High-level architectural overview of \eigent. (b) Detailed schematic of the Browser Agent, and added \qplus tools.}
  \Description{System architecture diagram showing (a) the high-level \eigent multi-agent framework and (b) the Browser Agent with additional \qplus tools for query processing, targeted extraction, and research progress reflection.}
  \label{fig:arch}
\end{figure}

\section{Related Work}
Deep research agents are typically equipped with tools that have external abilities, like web search APIs (e.g., CoSearchAgent~\cite{gong2024cosearchagentlightweightcollaborativesearch}, Agentic Reasoning~\cite{wu2025agenticreasoningstreamlinedframework}, OpenManus~\cite{openmanus}),
 browser tools (e.g., AutoAgent~\cite{tang2025autoagentfullyautomatedzerocodeframework}, DeepResearcher~\cite{zheng2025deepresearcherscalingdeepresearch}), and coding tools (e.g., Open Deep Research~\cite{huggingfaceopendeepresearch}). To our best knowledge, there are no existing open-source deep research agent projects using dedicated reasoning tools to guide the agents to perform structured reasoning. The reasoning process is typically implemented in a ReAct~\cite{yao2023reactsynergizingreasoningacting} style, where reasoning is free-style and \textit{interleaved} with actions (e.g., Search-o1, Search-R1, Agent-R1~\cite{cheng2025agentr1trainingpowerfulllm}, R1-Searcher~\cite{song2025r1searcherincentivizingsearchcapability}) or managed with fixed planning workflows (e.g., AvaTar~\cite{wu2024avataroptimizingllmagents}, The AI Scientist~\cite{lu2024aiscientistfullyautomated}, DeerFlow~\cite{deerflow}). Our \qplus tools provide a unique perspective, where dedicated reasoning tools explicitly structure the cognitive process for deep research agents.

Our reasoning approach is inspired by Anthropic's ``think tool'' paradigm~\cite{anthropic2025claude_think_tool}, and also incorporates the concepts of traditional information retrieval (IR), such as query generation, query expansion, and search frontier management~\cite{manning2008introduction}. In traditional systems, these are algorithmic processes (e.g., pseudo-relevance feedback~\cite{carpineto2012survey}). \qplus bridges this gap by mapping established IR principles onto the agent's tool space. Using dedicated reasoning tools for query processing and evidence aggregation, \qplus effectively translates classic IR strategies into structured, model-driven tool invocations.

\begin{figure*}[t]
  \centering
  \includegraphics[width=0.58\linewidth]{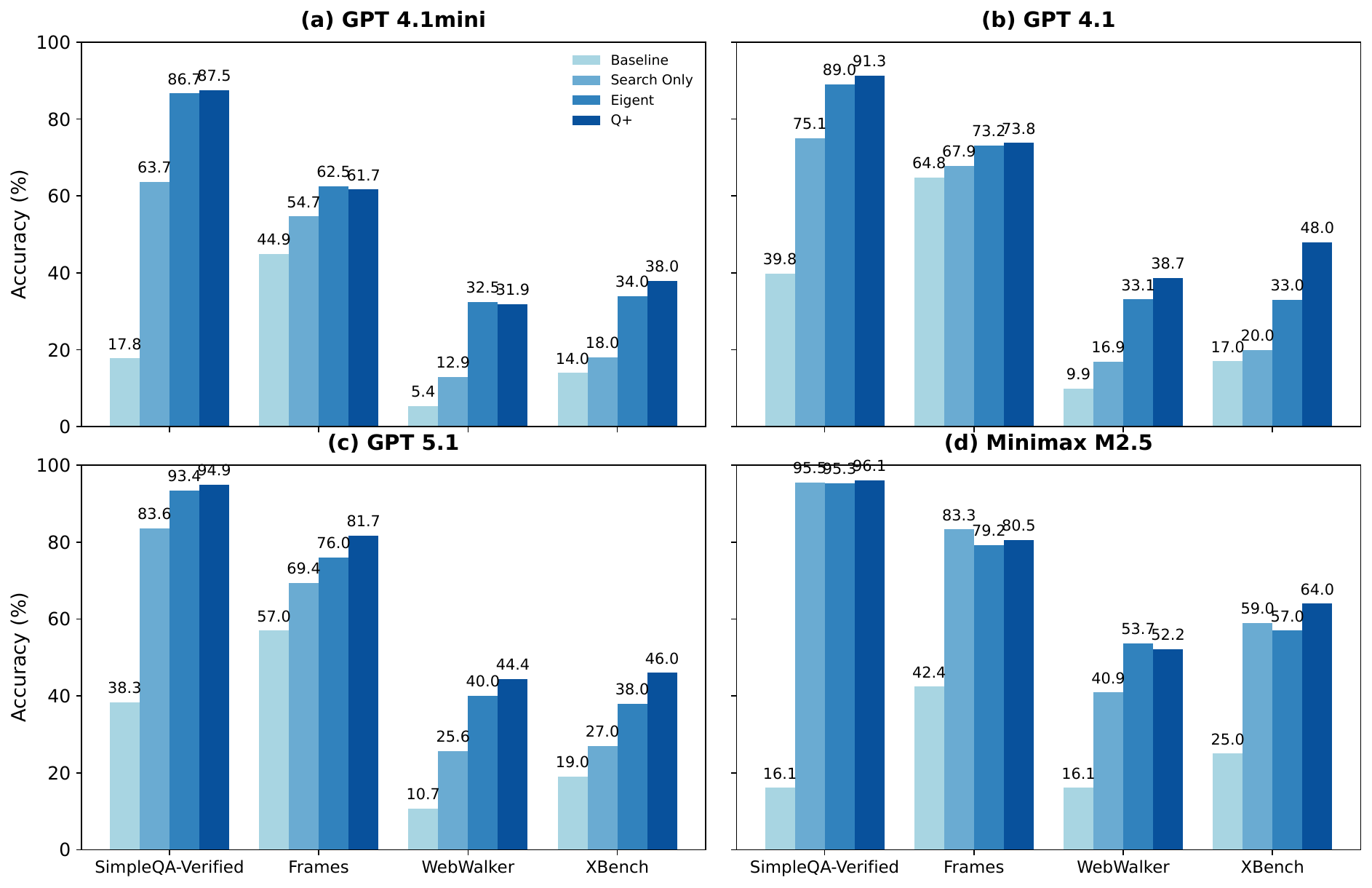}
  \caption{Performance analysis of agent configurations across four LLM backends. Accuracy results for \textbf{GPT-4.1 mini} (a), \textbf{GPT-4.1} (b), \textbf{GPT-5.1} (c), and \textbf{Minimax M2.5} (d) across four datasets. The \qplus framework is evaluated against Direct Generation (Baseline), Search Only, and \eigent browser agent.}
  \Description{A four-panel benchmark figure comparing accuracy across four datasets for GPT-4.1 mini, GPT-4.1, GPT-5.1, and Minimax M2.5. In most panels, \qplus outperforms the baseline, Search Only, and the standard \eigent browser agent, with the largest gains on stronger GPT backends.}
  \label{fig:perf}
\end{figure*}

\section{System Overview}

\eigentsearchqplus is an agent system based on \eigent's browser agent, enhanced with structured reasoning tools for query generation, information extraction, and research progress analysis.
\Cref{fig:arch} provides a schematic diagram of the relationship between \eigent, its browser agent, and the \qplus tools.
In this section, we first give a high-level overview of \eigent's architecture, then describe its browser agent in more detail, and finally introduce the \qplus tools.

\subsection{\eigent Overview}
\eigent builds on Camel-AI's workforce infrastructure~\cite{li2023camelcommunicativeagentsmind, hu2025owloptimizedworkforcelearning} (\Cref{fig:arch}(a)), and targets \emph{computer use} via a multi-agent workforce: the workforce engine dynamically decomposes high-level tasks into sub-tasks, activates multiple agents in parallel according to required capabilities, and coordinates their intermediate results. The system also supports human-in-the-loop intervention when tasks encounter uncertainty or require manual guidance. \eigent includes four specialized sub-agents:
\begin{itemize}
  \item \textbf{Browser Agent:} Performs web searches with search APIs and navigates websites using a browser toolkit~\cite{camelbrowsertoolkitblog}; it is also equipped with note-taking and terminal commands as auxiliary tools.\footnote{In the production \eigent system, the browser agent can request human input when it cannot resolve a query. We disable this capability in our experiments to evaluate autonomous performance.}
  \item \textbf{Developer Agent:} Executes code and terminal commands (e.g., Python/Bash), captures screenshots, supports lightweight web deployment, and asks for human feedback.
  \item \textbf{Multimodal Agent:} Analyzes video and audio, performs speech-to-text transcription, and generates images.
  \item \textbf{Document Agent:} Creates and edits common document formats (e.g., DOCX, PDF, spreadsheets, and slides) and supports cloud storage integration.
\end{itemize}

\subsection{Browser Agent}
The browser agent of \eigent is equipped with four toolkits:
\begin{enumerate}
  \item \textbf{Search toolkit:} Uses a search API to identify relevant resources. In our current study, we use \texttt{search\_google}, which is built on the Google Custom Search JSON API, as the entry point to the web-search process; the agent then optionally requests the browser toolkit to visit selected URLs for deeper navigation.
  \item \textbf{Browser toolkit:} A hybrid (Python/Typescript) browser toolkit~\cite{camelbrowsertoolkitblog} which supports common navigation functions such as opening/closing the browser, switching tabs, and moving forward/backward. For example, the \texttt{browser\_visit\_page} tool visits an agent-selected URL and returns the full snapshot, and the \texttt{browser\_get\_som\_screenshot} tool captures a screenshot with interactive elements highlighted and labeled with reference IDs.
  \item \textbf{Terminal toolkit:} Allows the agent to execute shell commands (e.g., \texttt{curl}) for auxiliary retrieval and inspection.
  \item \textbf{Note-taking toolkit:} Enables persistent reminders and intermediate information tracking during long-horizon search.
\end{enumerate}
\subsection{\eigentsearchqplus}
\qplus augments \eigent's browser agent with \emph{query-processing} and \emph{evidence-processing} tools that externalize intermediate decisions as tool calls.
We follow Anthropic's ``think''-tool paradigm~\cite{anthropic2025claude_think_tool}: a \emph{think tool} is a \emph{trace-only} self-tooling interface that records an intermediate reasoning artifact (e.g., a plan, decision, or evidence summary) without retrieving new external information.
All \qplus tools adopt this interface.

\paragraph{Query-processing tools.}
\eigent's browser agent directly issues web searches via the \texttt{search\_google} tool, without an explicit query reformulation module (e.g., rewriting/expansion) to systematically decompose the information need.
\qplus introduces two tools for query decomposition and selection:
\begin{itemize}
  \item \textbf{\texttt{plan\_next\_searches}:} Generates follow-up queries as candidates by identifying knowledge gaps and applying query rewriting, expansion, and decomposition.
  \item \textbf{\texttt{select\_query\_and\_search}:} Selects a query from the frontier and executes the search.\footnote{\qplus is additive to \eigent's browser agent, except that \texttt{select\_query\_and\_search} replaces \texttt{search\_google}.}
\end{itemize}

\paragraph{Search state management.}
We maintain two query sets as a system-level constraint: \emph{frontier} (generated but unsearched) and \emph{explored} (executed).
\texttt{plan\_next\_searches} populates the frontier as a soft suggestion for the following searches; executing a query moves it to \emph{explored}.
Re-searching explored queries is blocked and returns an error.

\paragraph{Evidence-processing tools.}
To handle long web snapshots and to make the stopping decision explicit, we add:
\begin{itemize}
  \item \textbf{\texttt{extract\_relevant\_details}:} Extracts question- or query-relevant details from long snapshots returned by the browser toolkit.
  \item \textbf{\texttt{analyze\_search\_progress}:} Assesses whether the accumulated evidence is sufficient to answer the original question.
\end{itemize}

\begin{table*}[t]
  \centering
  \begin{tabular}{lrrrrr}
    \toprule
     & \textbf{SimpleQA-Verified} & \textbf{FRAMES} & \textbf{WebWalkerQA} & \textbf{XBench} & \textbf{Weighted Avg.} \\
    \midrule
    \textbf{GPT-4.1 mini} & 0.8 & -0.8 & -0.6 & 4.0 & 0.1 \\
    \textbf{GPT-4.1}      & 2.3 & 0.6  & 5.6 & 12.0 & 3.0 \\
    \textbf{GPT-5.1}      & 1.5 & 5.7  & 4.4 & 8.0 & 3.8 \\
    \textbf{Minimax M2.5} & 0.8 & 1.3  & -1.5 & 7.0 & 0.6 \\
    \bottomrule
  \end{tabular}
\caption{Accuracy improvements of \eigentsearchqplus over the \eigent browser agent. \textbf{All values are reported in percentage points (pp)} and represent the accuracy change of \qplus relative to \eigent (negative values indicate decreases). The weighted average summarizes the overall performance gain across all evaluated benchmarks.}
  \label{tab:relacc}
\end{table*}
\section{Evaluations}
\subsection{Benchmark Datasets}
We evaluate \eigentsearchqplus on four open-source benchmarks:%
{\setlength{\topsep}{0pt}%
\setlength{\partopsep}{0pt}%
\setlength{\parsep}{0pt}%
\setlength{\itemsep}{0pt}%
\begin{itemize}
  \item \textbf{SimpleQA-Verified}~\cite{haas2025simpleqaverifiedreliablefactuality}: SimpleQA~\cite{wei2024measuringshortformfactualitylarge} is a benchmark for short, fact-seeking questions, adversarially collected against GPT-4 and designed for unambiguous grading with a single indisputable answer per question. SimpleQA-Verified is a refined variant that reduces annotation noise and ambiguity (1000 questions).
  \item \textbf{FRAMES}~\cite{frames2024}: An end-to-end RAG benchmark for factuality, retrieval, and reasoning, built around multi-hop questions that require synthesizing evidence from multiple sources (824 questions).\footnote{When implementing the evaluation code, we did not find an official judge prompt for FRAMES, so we reused the SimpleQA judge prompt for this benchmark.} 
  \item \textbf{WebWalkerQA}~\cite{wu2025webwalkerbenchmarkingllmsweb}: A web-traversal benchmark that evaluates whether agents can systematically navigate subpages, following multi-step link paths to gather high-quality evidence for answering questions (680 questions).
  \item \textbf{XBench (DeepSearch)}~\cite{chen2025xbenchtrackingagentsproductivity}: A search-focused benchmark in XBench's AGI Tracking suite that targets agents' capabilities in search and retrieval through curated deep-search tasks (100 questions).
\end{itemize}
}

\subsection{Results}
We compare four agent configurations across all benchmarks: (i) \textbf{Direct Generation (Baseline)}: the model answers without external tools; (ii) \textbf{Search Only}: the agent can only call \texttt{search\_google}; (iii) \textbf{\eigent's browser agent (denoted as \eigent for short)}; and (iv) \textbf{\qplus}: the enhanced \eigentsearchqplus system.
We evaluate four LLM backends: \textbf{GPT-4.1 mini}, \textbf{GPT-4.1}, \textbf{GPT-5.1}, and \textbf{Minimax M2.5}. We set temperature to 0 for all runs. Reasoning effort is disabled for \textbf{GPT-5.1} but naturally enabled for \textbf{Minimax M2.5}. We use \textbf{GPT-4.1} as the automated judge across all experiments.
\Cref{fig:perf} summarizes accuracy across datasets and configurations.
Across the GPT-series models (\Cref{fig:perf}(a--c)), \eigent consistently outperforms \textsf{Search Only}, which in turn outperforms the \textsf{Baseline}. For \textbf{Minimax M2.5} (\Cref{fig:perf}(d)), \textsf{Search Only} performs comparably to or better than \eigent on SimpleQA-Verified and FRAMES, suggesting that the stronger base model can partially compensate for missing higher-level agent tooling. Overall, performance scales with model capability for the GPT series: \textbf{GPT-5.1} generally outperforms \textbf{GPT-4.1}, which outperforms \textbf{GPT-4.1 mini}.

\subsubsection{Comparative analysis: \eigent vs.\ \qplus}
The performance distinction between \qplus and the standard \eigent browser agent is more nuanced, as summarized in Table~\ref{tab:relacc}. For \textbf{GPT-4.1 mini}, the two configurations are broadly comparable, with variation across benchmarks. \qplus shows a notable gain on XBench; however, this result should be interpreted with caution given the benchmark's modest size (100 questions).

In contrast, for \textbf{GPT-4.1} and \textbf{GPT-5.1}, \qplus improves performance consistently across benchmarks, suggesting that the benefits of \qplus scale with base-model capability for GPT models (e.g., instruction following and multi-step reasoning).

For \textbf{Minimax M2.5}, the gap between \qplus and the \eigent agent is smaller. While \qplus underperforms on WebWalkerQA, it improves accuracy on all the other benchmarks, yielding a modest average gain. This finding is particularly notable given that \textbf{Minimax M2.5} is natively trained to perform interleaved, internal reasoning prior to action execution. The continued effectiveness of \qplus on this backend suggests a critical insight for compound AI systems: explicit, tool-enforced structured reasoning (system-level scaffolding) is complementary to the base model's internal cognitive capabilities (model-level reasoning). Even when an LLM is highly capable of implicit reasoning, forcing it to manage explicit IR states (such as tracking an explored frontier or isolating relevant evidence) yields reasonable gains.

Accuracy is important, but cost also matters. We find that \qplus consumes about 1.4$\times$ as many tokens as \eigent on average; we report this trade-off for transparency, and provide details in the Supplementary Material, Section ``Cost Analysis.''

\subsubsection{Case studies}
To better understand why \qplus outperforms the \eigent browser agent, we examined tool-calling trajectories of representative queries where \eigent failed but \qplus succeeded; see Supplementary Material, Section ``Case Studies.'' Across these cases, \qplus makes intermediate sub-questions, evidence quality checks, and extraction steps explicit, which yields more structured search behavior.
Qualitatively, \qplus helps avoid common failure modes in \eigent's browser agent, including under-decomposition of multi-step questions, reliance on high-level or outdated pages, and distraction by information-dense snapshots. These examples suggest that explicit query planning, progress monitoring, and targeted extraction together produce more reliable search trajectories.

\section{Conclusion}
We presented \eigentsearchqplus, an enhanced variant of \eigent's browser agent that augments it with lightweight, specialized modules for structured query processing and explicit search-time reasoning. Across multiple benchmarks, these additions yield overall average accuracy gains, while preserving the underlying agent design.

A key property of \qplus is its \emph{non-invasive} and modular architecture: the tools operate as optional add-ons that can be enabled selectively, requiring minimal changes to the core agent. This design makes \qplus broadly applicable as a reusable component for other deep-research agents, particularly in settings that require iterative query generation and selection, targeted evidence extraction from long web snapshots, and ongoing assessment of search progress and evidence sufficiency.

More broadly, our results suggest that exposing query-processing and evidence processing operations as explicit, inspectable tools can improve both the robustness and the interpretability of deep-research agent behavior. We hope this work encourages practitioners to experiment with such reasoning layers in agentic workflows to better understand their role in complex, multi-step information-seeking tasks.

\section{Limitations and Future Work}
As reasoning-oriented LLMs become more common, controlled comparisons between reasoning and non-reasoning backends are needed to clarify whether \qplus's explicit, tool-based structured reasoning complements or duplicates internal model reasoning. While we include Minimax M2.5, which is a reasoning model by design, broader controlled comparisons across reasoning-enabled and reasoning-disabled settings remain future work. Meanwhile, our current system is training-free; we are exploring fine-tuning and RL-based variants to test whether learning can further amplify \qplus's gains.

\newpage
\begin{acks}
We thank the three anonymous reviewers for their helpful feedback. This work was funded by CAMEL-AI.org and Eigent.ai. All authors contributed to this work as open-source contributors at CAMEL-AI.org. We thank Douglas Yueming Lai of CAMEL-AI.org for designing Figure~1, which illustrates the system architecture.

\end{acks}

\bibliographystyle{ACM-Reference-Format}
\bibliography{ref}

\end{document}